\documentclass[sigconf]{acmart}

\copyrightyear{2023}
\acmYear{2023}
\setcopyright{rightsretained}
\acmConference[SIGIR '23]{Proceedings of the 46th International ACM SIGIR Conference on Research and Development in Information Retrieval}{July 23--27, 2023}{Taipei, Taiwan}
\acmBooktitle{Proceedings of the 46th International ACM SIGIR Conference on Research and Development in Information Retrieval (SIGIR '23), July 23--27, 2023, Taipei, Taiwan}\acmDOI{10.1145/3539618.3592092}
\acmISBN{978-1-4503-9408-6/23/07}

\makeatletter
\gdef\@copyrightpermission{
 \begin{minipage}{0.3\columnwidth}
  \href{https://creativecommons.org/licenses/by/4.0/}{\includegraphics[width=0.90\textwidth]{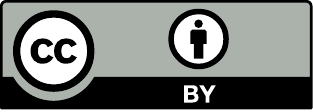}}
 \end{minipage}\hfill
 \begin{minipage}{0.7\columnwidth}
  \href{https://creativecommons.org/licenses/by/4.0/}{This work is licensed under a Creative Commons Attribution International 4.0 License.}
 \end{minipage}
 \vspace{5pt}
}
\makeatother

\usepackage{booktabs}
\usepackage{multirow}
\usepackage[normalem]{ulem}
\useunder{\uline}{\ul}{}
\usepackage{todonotes}
\begin{document}

\title{EmoUS: Simulating User Emotions in Task-Oriented Dialogues}


\author{Hsien-Chin Lin}
\email{linh@hhu.de}
\orcid{0009-0006-0027-226X} 
\author{Shutong Feng}
\orcid{0000-0002-1307-4223} 
\author{Christian Geishauser}
\orcid{0009-0004-6369-3438} 
\affiliation{%
  \institution{Heinrich Heine University Düsseldorf}
  \streetaddress{Universitätsstraße 1}
  \city{Düsseldorf}
  \country{Germany}
  \postcode{40225}
}
\author{Nurul Lubis}
\orcid{0000-0002-4461-7243} 
\author{Carel van Niekerk}
\orcid{0000-0002-4551-7447} 
\author{Michael Heck}
\orcid{0000-0001-9841-5025} 
\affiliation{%
  \institution{Heinrich Heine University Düsseldorf}
  \streetaddress{Universitätsstraße 1}
  \city{Düsseldorf}
  \country{Germany}
  \postcode{40225}
}
\author{Benjamin Ruppik}
\orcid{0000-0001-9035-9217} 
\author{Renato Vukovic}
\orcid{0000-0001-6303-9402}
\author{Milica {Gašić}}
\email{gasic@hhu.de}
\orcid{0000-0003-0318-9147} 
\affiliation{%
  \institution{Heinrich Heine University Düsseldorf}
  \streetaddress{Universitätsstraße 1}
  \city{Düsseldorf}
  \country{Germany}
  \postcode{40225}
}
\renewcommand{\shortauthors}{Hsien-Chin Lin et al.}


\begin{abstract}
Existing user simulators (USs) for task-oriented dialogue systems only model user behaviour on semantic and natural language levels without considering the user persona and emotions. Optimising dialogue systems with generic user policies, which cannot model diverse user behaviour driven by different emotional states, may result in a high drop-off rate when deployed in the real world. Thus, we present EmoUS, a user simulator that learns to simulate user emotions alongside user behaviour. EmoUS generates user emotions, semantic actions, and natural language responses based on the user goal, the dialogue history, and the user persona. By analysing what kind of system behaviour elicits what kind of user emotions, we show that EmoUS can be used as a probe to evaluate a variety of dialogue systems and in particular their effect on the user's emotional state. Developing such methods is important in the age of large language model chat-bots and rising ethical concerns.
\end{abstract}



\maketitle
\section{Introduction}
Task-oriented dialogue systems (DSs) help users accomplish their goals, such as searching for nearby restaurants or booking a hotel. 
Proficient DSs are often trained via reinforcement learning (RL), which demands a large number of interactions between the system and users, making training with real users expensive and time-consuming. Therefore, utilizing user simulators (USs) to build a controlled interactive environment becomes attractive \cite{658991}.

Despite recent USs in task-oriented dialogues properly modelling user extrinsic behaviour in terms of semantic actions and natural language \cite{lin-etal-2022-gentus, tseng2021transferable}, a crucial aspect is still lacking: the user intrinsic state such as user persona and the emotional state. 
A generic user policy may lead to limited linguistic diversity and fails to capture diverse actions driven by varying user emotions.
Adjusting the probability distribution of user actions in rule-based USs is a popular method to address diversity \cite{keizer-etal-2010-parameter}, but real users differ in more ways than just action preferences. 
Training USs by supervised learning with different initialisation \cite{tang-etal-2021-high} or by RL with varying reward functions can also form various user policies \cite{lin-etal-2022-gentus}, but that can only provide diverse extrinsic behaviour, e.g. the action length in each turn or the semantic content.

In this work, we propose a user simulator that models the user emotional state conditioned on the dialogue context and the user persona. More specifically, our contributions are as follows:
\begin{itemize}
    \item We propose an \textbf{emo}tional \textbf{u}ser \textbf{s}imulator that we call \emph{EmoUS}\footnote{\url{https://gitlab.cs.uni-duesseldorf.de/general/dsml/emous-public}}. 
    The \emph{EmoUS} response includes the user emotion, semantic actions, and natural language utterances. To the best of our knowledge, this is the first user simulator with user emotion for task-oriented dialogue systems.
    \item EmoUS exhibits an increased linguistic diversity for the same context by modelling the user policy and emotion jointly, 
    \item The user emotion of EmoUS provides valuable insights for evaluating DSs, offering a more subtle and detailed understanding beyond a simple measure of task success.
\end{itemize}
\section{Related work}
\label{sec:related}

The effectiveness of a task-oriented dialogue policy trained by RL with a US is greatly affected by the quality of the US~\cite{schatzmann2005quantitative}. 
Rule-based USs are commonly used to train DSs, such as the agenda-based US (ABUS)~\cite{schatzmann-etal-2007-agenda}. ABUS models the user goal as a stack-like agenda, ordered by the priority of the user actions updated by hand-crafted stacking and popping rules. 
While its action probability distribution can be manipulated to simulate different user behaviour~\cite{keizer-etal-2010-parameter}, it only generates semantic actions without natural language generation or emotion prediction. 
Moreover, designing rules for complex scenarios is labour-intensive and transferring these rules to new domains can be challenging. 
To address these limitations, data-driven USs have been developed, which learn user policy directly from data. The sequence-to-sequence (Seq2Seq) model structure is the most common framework. The input sequence may include the dialogue history and user goal as a list of features or plain text. The output sequence can be semantic actions or natural language utterances \cite{el2016sequence, kreyssig-etal-2018-neural, gur2018user, lin-etal-2021-domain, tseng-etal-2021-transferable, wan-etal-2022-unified, lin-etal-2022-gentus}. 
\citet{tang-etal-2021-high} train USs by supervised learning with different initialisation to create different user policies. \citet{lin-etal-2022-gentus} proposed GenTUS, an ontology-independent US which generates natural language utterances as well as the underlying semantic actions for interpretability. Its behaviour can be configured by RL with different reward functions.
These USs can simulate extrinsic user behaviour, e.g. actions and utterances, but intrinsic user states are neglected, e.g. satisfaction level and emotional status \cite{poria2019emotion}.

In comparison to generating responses with given emotions~\cite{song-etal-2019-generating, 10.1007/s10489-021-02819-z, colombo-etal-2019-affect} or recognising user satisfaction classification after receiving user utterances~\cite{engelbrecht-etal-2009-modeling, hara-etal-2010-estimation, higashinaka-etal-2010-modeling, schmitt2015interaction, bodigutla2019multi, 9746464}, the user satisfaction modelling should predict intrinsic user states first then generates actions or utterances.  
\citet{sun2021simulating} and \citet{10.1145/3485447.3512020} investigate how user satisfaction impacts user behaviour on the semantic level. 
\citet{pan-etal-2022-user} transfer the emotion from chit-chat to task-oriented dialogues utilising data augmentation. 
\citet{kim2022multi} proposed SatActUtt, which generates users' satisfaction, action (only with intent and domain), and utterance based on dialogue history as multi-task learning. We consider SatActUtt as our baseline as it is the first US modelling both intrinsic and extrinsic user behaviour. 
While SatActUtt can predict user satisfaction scores adequately based on dialogue history, it does not include the user goal. This makes it difficult to train a dialogue system. In addition, it only considers satisfaction and dissatisfaction, disregarding aspects such as different emotion elicitors or user personas~\cite{ortony_clore_collins_1988, lubis2016emotion}. \citet{feng-etal-2022-emowoz} annotated a task-oriented dialogue dataset with 7 user emotions inspired by Ortony, Clore and Collins (OCC) emotion model \cite{ortony_clore_collins_1988}. It considers user conduct and emotion elicitors for human-human and human-machine task-oriented dialogues, representing a more fine-grained user intrinsic state for task-oriented dialogues.

\section{Simulating user emotion in task-oriented dialogues}
\label{sec:emous}
Task-oriented DSs are underpinned by an \emph{ontology} which is typically composed of \emph{intents}, \emph{domains}, \emph{slots}, and \emph{values}. \emph{Intents} define user or system global intentions of their respective actions in each turn. Users and systems may have different intents, e.g., systems can \emph{confirm} user's request and users can \emph{negate} system's proposal. \emph{Domains} are the topics that can be discussed in the conversation. They can be further specified by \emph{slots} and each can take a number of \emph{values}. We assume that the users of task-oriented dialogues have a \emph{goal} they want to achieve, which is defined as $G = \{ d_1: [ (s_1, v_1), (s_2, v_2), \dots], d_2: [ (s_3, v_3),  \dots], \dots\}$,
where domain $d_i$, slot $s_i$ and value $v_i$ are selected from the ontology.

Semantic \emph{user actions} and \emph{system actions} are composed of tuples, $(intent, domain, slot, value)$.  Semantic actions can be transformed into natural language utterances.
User \emph{emotion} in task-oriented dialogues may be triggered by different elicitors, or related to different user personas. For example, the system not responding adequately may lead to the user being dissatisfied ~\cite{feng-etal-2022-emowoz}. A user \emph{persona} represents users' attitudes and feelings towards certain events, such as feeling fearful after a robbery \cite{mairesse-walker-2006-automatic} or includes users' conduct, i.e.\ how users behave when communicating, e.g. are they polite or impolite. For example, the persona of a polite user who is feeling excited to visit a museum is $persona = \text{\{user: polite, attraction: excited\}}$. The user persona can be derived from dialogue history during training and sampled from a distribution for inference. 

User simulation with emotion can be viewed as a Seq2Seq problem. For each turn, EmoUS predicts the user emotion based on the context information, e.g. the dialogue history, the user goal, and the user persona, and generates semantic actions and natural language responses based as follows. 

\subsection{Model structure}
\begin{figure}[th]
    \centering
    \includegraphics[width=\columnwidth]{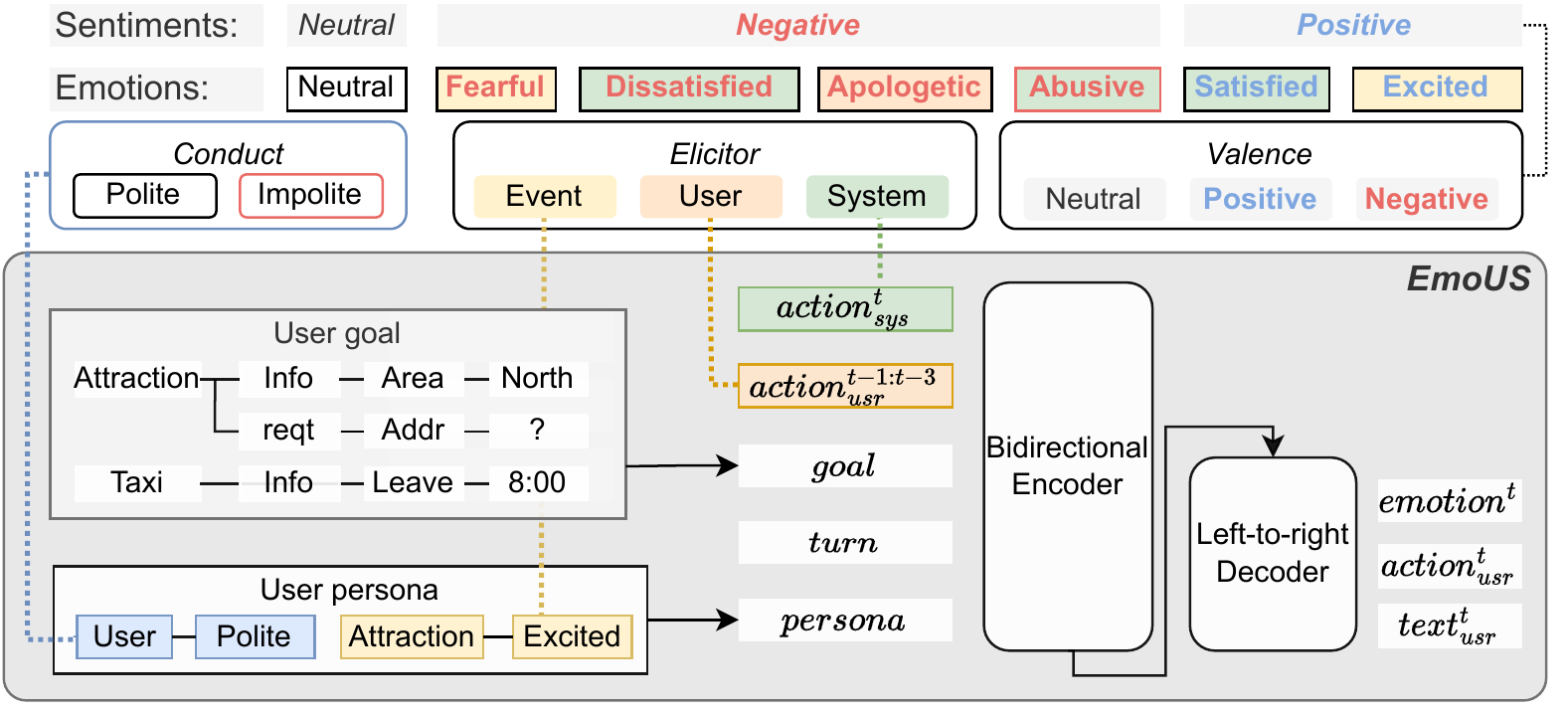}
    \caption{The model structure of EmoUS}
    \label{fig:structure}
\end{figure}
EmoUS builds upon GenTUS and additionally incorporates user persona and emotion prediction. More specifically, 
EmoUS takes the system action $action^t_{sys}$, user history $action^{t-1:t-3}_{usr}$, user goal $goal$, turn information $turn$ and the user persona $persona$ as input and generates user emotion $emotion$, semantic actions $action^t_{usr}$, and an utterance $text^t_{usr}$ as output at turn $t$ (as shown in Fig.~\ref{fig:structure}).
By introducing different user personas and emotions, more diverse user behaviours on both semantic and linguistic aspects can be simulated even in the same context.

EmoUS considers the three aspects of user emotions in task-oriented dialogues according to EmoWOZ \cite{feng-etal-2022-emowoz}, namely \emph{elicitor}, \emph{conduct}, and \emph{valence}. The emotion elicitor can be an event, the system, or the user. Their respective information can be captured from the event $persona$ attribute, system $action^t_{sys}$, and user $action^{t-1:t-3}_{usr}$. The user conduct, whether polite or impolite, is recorded as a user persona. The valence aspect, or the sentiment polarity of each emotion, is informed intrinsically in the emotion prediction.

%

Following the setting in \citet{lin-etal-2022-gentus}, the input and output sequences are represented as JSON-formatted strings, 
composed of natural language tokens. In this way, EmoUS achieves ontology independence and can transfer to unseen domains.\footnote{As this property is directly inherited from GenTUS, we do not examine it in our experiments.}
Then we train EmoUS as a Seq2Seq model and leverage BART~\cite{lewis2020bart}, a transformer-based natural language generator with a bidirectional encoder and a left-to-right decoder. BART demonstrates impressive performance in a range of language-related tasks.

\section{Experimental Setup}
\label{sec:set-up}
The aim of our experiments is to demonstrate that EmoUS is able to generate user emotion, semantic actions, and utterances based on the context of the conversation and the user persona. Furthermore, we show that the emotion prediction of EmoUS is a valuable tool for evaluating DSs. 
We conduct our experiments on EmoWOZ \cite{feng-etal-2022-emowoz}. 
It contains user emotion annotations for human-human dialogues from MultiWOZ \cite{budzianowski-etal-2018-multiwoz} and $1k$ human-machine dialogues between volunteers and an RNN-based dialogue policy trained on MultiWOZ.

\subsection{Supervised learning for emotion simulation}
Our model is inherited from Huggingface's transformers \citep{wolf-etal-2020-transformers} and trained on EmoWOZ. To measure the emotion prediction performance, we calculate the macro-F1 score of sentiments and emotions. We compare sentiment prediction against SatActUtt \cite{kim2022multi}, a user model which predicts sentiment, user action (composed with intent and domain only), and utterances based on the dialogue history.

Following the setting of \citet{lin-etal-2022-gentus}, we evaluate the performance of modelling user semantic actions by F1 score and turn accuracy and the natural language generation (NLG) performance by slot error rate (SER), sacre-BLEU score \citep{post-2018-call} and self-BLEU score \citep{zhu2018texygen}. 
SER measures the agreement between the semantic actions and the corresponding utterance. $SER = (m+h)/N$, where $N$ is the total number of slots in semantic actions, $m$ and $h$ stand for the number of missing and hallucinated slots. 
The self-BLEU evaluates the diversity of generated utterances in the following way. 
After generating a sentence for every data point, we calculate a BLEU score by treating all other generated sentences as references. By averaging these scores, we get the self-BLEU score where the lower score implies a higher diversity.

\subsection{Interacting with DS}
We estimate the generalisation ability of a US by cross-model evaluation, where a DS trained with a particular US is evaluated by different USs \cite{schatztnann2005effects}. Policies of different DSs are trained with various USs, including the agenda-based US (ABUS) with T5 \cite{raffel2020exploring} natural language generator (ABUS-T5), GenTUS, and EmoUS,  by proximal policy optimisation (PPO) \cite{schulman2017proximal}, a simple and stable RL algorithm, for $200$ epochs, each of which consists of $1000$ dialogue turns.  
Each policy is trained on $5$ random seeds and the performance is averaged.
The DSs also include a natural language understanding module composed with BERT \citep{devlin-etal-2019-bert} for understanding users' utterances and a rule-based dialogue state tracker for tracking the users' states under the ConvLab-3 framework \cite{zhu2022convlab}.

We also analyse how different system behaviour elicit user emotions. For this purpose, we used $1k$ dialogues between EmoUS and a DS trained by EmoUS. We categorised various system behaviour in the following groups: 
\emph{confirm} - the system repeats the slots and values informed by the user, \emph{no\_confirm} - the system does not repeat this information, \emph{miss\_info} - the system requests the information just mentioned by the user, \emph{neglect} - the system does not respond to the user request, \emph{reply} - the system responds to the user request, and \emph{loop} - the system takes identical actions for two turns in a row. 
\section{Experimental Results}
\label{sec:results}
\subsection{User emotion modelling}
As shown in Table~\ref{tab:emotion-model}, EmoUS outperforms SatActUtt on sentiment prediction by $0.314$ on macro-F1 score. This is not unexpected as EmoUS includes the user goal in inputs and the user sentiment in task-oriented dialogues is centred around the user goal \cite{feng-etal-2022-emowoz}. In addition, the performance of sentiment prediction between EmoUS and $\text{EmoUS}_{noPersona}$ is similar, but the emotion prediction improves by $0.202$ on the macro-F1 score when including the user persona. This indicates that considering the user persona improves the performance of user emotions triggered by different elicitors.

\begin{table}[h]
\centering
\caption{Performance for emotion and sentiment prediction of different models by measuring macro-F1 score.}
\begin{tabular}{@{}lcc@{}}
\toprule
\textbf{model} & \textbf{sentiment} & \textbf{emotion} \\ \midrule
SatActUtt & 0.379 & - \\
$\text{EmoUS}_{noPersona}$ & 0.673 & 0.299 \\
EmoUS & \textbf{0.693} & \textbf{0.501} \\ \bottomrule
\end{tabular}
\label{tab:emotion-model}
\end{table}

\begin{figure}[h]
    \centering
    \includegraphics[width=\columnwidth]{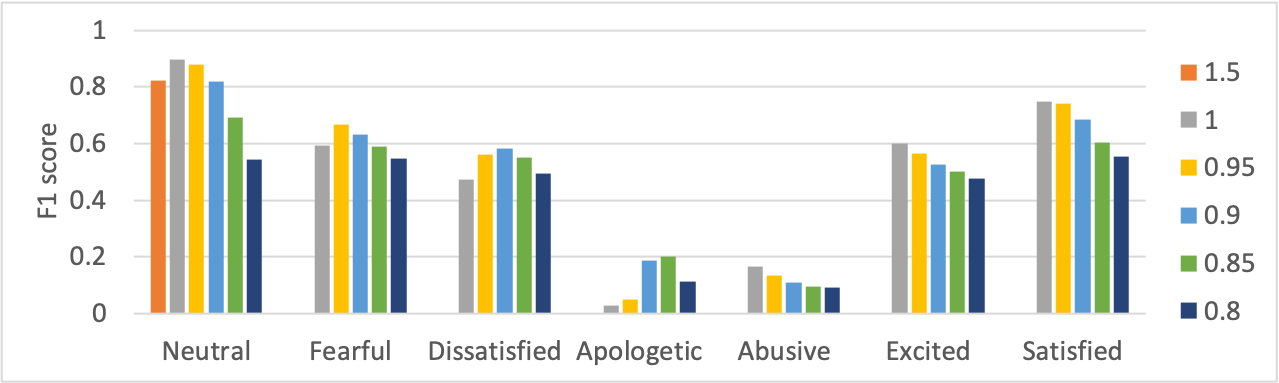}
    \caption{Different weights of the neutral emotion will have different F1-score on each user emotion.}
    \label{fig:configure_on_weight}
\end{figure}

We demonstrate that user emotion simulation can be further configured by multiplying different weights $w$ on the probability of \emph{neutral}, i.e. \emph{neutral} is more likely to be selected with a higher weight. As shown in Fig.~\ref{fig:configure_on_weight}, EmoUS is purely neutral without any emotion as $w=1.5$. As the weight decreases, EmoUS achieves the best performance on \emph{fearful} as $w=0.95$, best on \emph{dissatisfied} as $w=0.9$, and best on \emph{apologetic} as $w=0.85$. Thus, we can change the probability distribution of emotions in the user response, inducing different user behaviour, by modifying the weight of emotions.

\subsection{User action prediction}
The results of user action prediction are shown in Table~\ref{tab:action-model}, where $\text{EmoUS}_{emo}$ generates semantic actions based on golden emotions. 
EmoUS is superior to SatActUtt because EmoUS can generate semantic actions following the user goal, whereas SatActUtt does not consider the user goal. Additionally, EmoUS is still comparable to GenTUS despite it models a more complex task, simulating user emotions and semantic actions jointly. Moreover, $\text{EmoUS}_{emo}$ surpasses GenTUS since $\text{EmoUS}_{emo}$ generates semantic actions utilising more information then GenTUS, such as the user persona and golden emotions. 

\begin{table}[h]
\centering
\caption{Performance for user action prediction.}
\label{tab:action-model}
\begin{tabular}{@{}lcccc@{}}
\toprule
 & \multicolumn{2}{c}{\textbf{Intents+domains}} & \multicolumn{2}{c}{\textbf{Full action}} \\
\textbf{model} & F1 & ACC & F1 & ACC \\ \midrule
GenTUS & 0.890 & 0.854 & 0.762 & 0.600 \\
SatActUtt & 0.317 & 0.221 & - & - \\
EmoUS & 0.892 & 0.857 & 0.764 & 0.603 \\
$\text{EmoUS}_{emo}$ & \textbf{0.904} & \textbf{0.867} & \textbf{0.775} & \textbf{0.611} \\ \bottomrule
\end{tabular}
\end{table}

\subsection{Natural language evaluation}
The NLG results are shown in Table~\ref{tab:NLG-model}, where $\text{GenTUS}_{act}$ generates utterances based on golden semantic actions and $\text{EmoUS}_{emo+act}$ is based on golden emotion and semantic actions. On the other hand, GenTUS and EmoUS are generated based on their prediction. The Sacre-BLEU is calculated with golden utterances.

Although SatActUtt generates the most diverse utterances with the lowest Self-BLEU score, it also has the lowest Sacre-BLEU score, which means it by and large generates random responses irrelevant to the user goal. On the other hand, $\text{EmoUS}_{emo+act}$ has a comparable Sacre-BLEU and SER with $\text{GenTUS}_{act}$ and a much lower Self-BLEU score, which means EmoUS is able to generate more diverse responses than GenTUS but still follows the user goal and maintains the agreement between the semantics and the language.

\begin{table}[h]
\centering
\caption{The NLG performance on EmoWOZ of different models. The arrow directions represent which trend is better.}
\begin{tabular}{@{}lccc@{}}
\toprule
\textbf{model} & \textbf{SER↓} & \textbf{Sacre-BLEU↑} & \textbf{Self-BLEU↓} \\ \midrule
Human & 0.054 & - & 0.770 \\
GenTUS & 0.116 & - & 0.950 \\
$\text{GenTUS}_{act}$ & \textbf{0.092} & \bf19.61 & 0.930 \\
SatActUtt & - & 2.90 &  \bf0.433\\
EmoUS & 0.118 & - & 0.715 \\
$\text{EmoUS}_{emo+act}$ & 0.096 & 16.91 & 0.708 \\ \bottomrule
\end{tabular}

\label{tab:NLG-model}
\end{table}

\subsection{Cross-model evaluation}
As shown in Table~\ref{tab:cross-model}, the DS trained with EmoUS performs comparably to the DS trained with ABUS-T5 when evaluating with ABUS-T5 ($0.62$ vs $0.63$ success rate), and similarly to the DS trained with GenTUS when evaluating with GenTUS (both at $0.53$ success rate). However, the DS trained with EmoUS outperforms the DS trained with ABUS-T5 by 7\% absolute and the DS trained with GenTUS 5\% absolute on success rate when evaluating with EmoUS (success rates of $0.52$ vs $0.45$ and $0.47$ respectively). This indicates that EmoUS provides a better learning environment and makes DSs trained with it perform well when evaluated on diverse USs.

\begin{table}[ht]
\centering
\caption{The success rates of policies trained on EmoUS, GenTUS, and ABUS with T5 NLG (ABUS-T5) when tested on various USs. Each pair is evaluated by 400 dialogues on 5 seeds, which is 2K dialogues in total.}
\begin{tabular}{@{}lccc@{}}
\toprule
\textbf{US for} & \multicolumn{3}{c}{\textbf{US for evaluation}} \\
\textbf{training} & ABUS-T5 & GenTUS & EmoUS \\ \midrule
ABUS-T5 & 0.63 & 0.48 & 0.45 \\
GenTUS & 0.60 & 0.53 & 0.47 \\
EmoUS & 0.62 & 0.53 & \textbf{0.52} \\ \bottomrule
\end{tabular}

\label{tab:cross-model}
\end{table}

\subsection{System behaviour with the user emotions}
In $1k$ dialogues between EmoUS and a DS trained by it, the system behaviour $confirm$, $no\_confirm$, and $miss\_info$ elicit neutral emotion.
As systems respond properly, e.g. $reply$, users are likely to feel satisfied, but when systems behave unprofessionally, e.g. $neglect$ and $loop$, users may feel dissatisfied (see Table~\ref{tab:system_act}). This observation is in line with the user study conducted by \citet{sun2021simulating}.

\begin{figure}[h]
    \centering
    \includegraphics[width=\columnwidth]{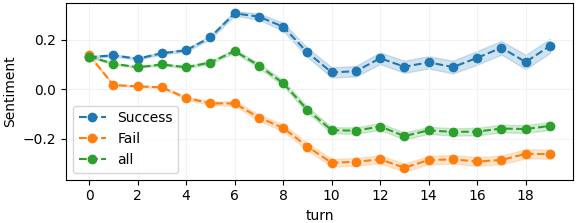}
    \caption{The average user sentiment in different turns.}
    \label{fig:sentiment_turn}
\end{figure}

Furthermore, we plot the average user sentiment per turn where $\text{positive} =+1$, $\text{neutral}=0$, and $\text{negative}=-1$, for each dialogue outcome.
As expected, users are more positive in successful dialogues and more negative in failed dialogues on average~(see Fig.~\ref{fig:sentiment_turn}). In addition, we also notice a drop in sentiment around turn 6, which suggests the user may feel impatience after that. 
\begin{table}[h]
\centering
\caption{Proportion of neutral and system-eliciting emotions triggered by various system behaviour.}
\label{tab:system_act}
\begin{tabular}{@{}lrrr@{}}
\toprule
\bf{System} & \multicolumn{3}{c}{\bf User emotion} \\
\bf{behaviour} & Neutral & Dissatisfied & Satisfied \\ \midrule
confirm & 86.00\% & 2.20\% & 11.80\% \\
no\_confirm & 71.80\% & 16.60\% & 11.60\% \\
miss\_info & 79.20\% & 11.10\% & 9.70\% \\
neglect & 27.10\% & 65.00\% & 7.90\% \\
reply & 51.50\% & 4.10\% & 44.40\% \\
loop & 28.60\% & 65.90\% & 5.50\% \\ \bottomrule
\end{tabular}
\end{table}


\section{Conclusion}
\label{sec:conclusion}
 We present EmoUS, a simulated user that generates emotional and thus more diverse output in task-oriented dialogues. 
 It can be further configured by manipulating different weights for each emotion or different user personas. Our results show that EmoUS is useful to examine the impact of dialogue systems on the user's emotional state. Developing such probes is of particular importance with the increasing usage of dialogue systems and the rising ethical concerns of large language model chat-bots. 
 
 In future, 
 the correlations between personas and emotions should be investigated, e.g. whether polite users show more satisfaction even though system responses are inadequate. Human evaluation should also be conducted to address the validity of our simulation. Furthermore, we plan to utilise EmoUS for the development of emotion-sensitive DSs.

\bibliographystyle{ACM-Reference-Format}
\balance
\bibliography{custom}
\end{document}